# SVD-EBP Algorithm for Iris Patten Recognition

Mr. Babasaheb G. Patil
Department of Electronics Engineering
Walchand College of Engineering,
Sangli, India
patilbg@rediffmail.com

Dr. Mrs. Shaila Subbaraman
Department of Electronics Engineering
Walchand College of Engineering,
Sangli, India
shailasubbaraman@yahoo.co.in

*Abstract—* **This paper proposes a neural network approach based on Error Back Propagation (EBP) for classification of different eye images. To reduce the complexity of layered neural network the dimensions of input vectors are optimized using Singular Value Decomposition (SVD). The main of this work is to provide for best method for feature extraction and classification. The details of this combined system named as SVD-EBP system, and results thereof are presented in this paper.**

Keywords- *Singular value decomposition(SVD)*, Error back Propagation(EBP).

## I. INTRODUCTION

Biometrics is one of the areas of research that has gained widespread acceptance as form of unique human identification and fraud prevention. Although the current state-of-the-art provides reliable automatic recognition of biometric features, the field is not completely researched. Different biometric features offer different degrees of reliability and performance. The Human Iris is one of the best biometrics feature in the human body for person recognition. This paper provides a walkthrough for image acquisition, image segmentation, feature extraction and pattern forming based on the Human Iris imaging. A Feed forward Neural Network is implemented for classifying the various Iris patterns and then verifying one's identity.

### A. Characteristics of Human Iris

The use of the Human Iris as a biometric feature offers many advantages over other human biometric features. The Iris is the only internal human body organ that is visible from the outside and is well protected from external modifiers. A fingerprint for example may suffer transformations due to harm or aging, voice patterns may be altered due to vocal diseases. However, the human Iris image is relatively simple to acquire and may be done so in a non-intrusive way. The Human Iris starts forming right from the third month of gestation in the mother's uterus. A small part of final Iris pattern is developed from the individual DNA while most of the part is developed randomly by the growth of epithelial tissues present there. It means that two eyes from the same individual, although they look very similar, have two different patterns of two Irises, however with unique DNA related internal pattern. Identical twins would then exhibit four different Iris patterns..

## II. PREPROCESSING

This paper consists of two parts; the first part describes the preprocessing techniques while, the second part uses data generated by the work of first part for pattern classification using a Feed Forward Neural Network. Fig. 1 shows the steps involved in this paper.

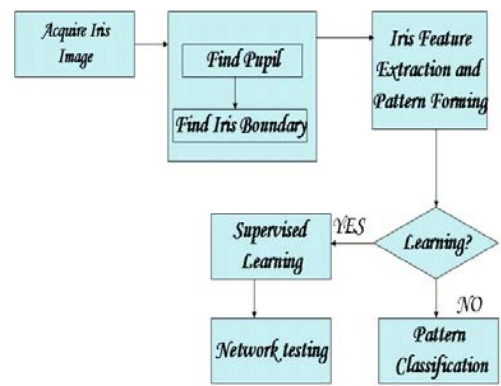

Fig.1: Processing Steps

### A. *Image Acquisition*

The major problem of Iris Recognition is image acquisition because of the susceptibility of eyes to degree of illumination. It is very important that one system implements consistent illumination. Also, the pupil is an open door to the retina, one of the most sensitive organs of our body, and extra care must be taken when shedding direct light over it.

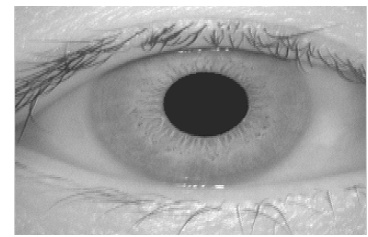

Fig .2 : Image of an Eye

The work presented in this paper uses the CASIA Iris database as input. This database uses a special camera that operates in the infrared spectrum of light, not visible by the human eye. Here, each Iris class is composed of 7 samples taken in two sessions, three in the first session and four in the second session. The two sessions were taken with an interval of one month. Images are 320x280 pixels gray scale taken by a digital optical sensor designed by NLPR (National Laboratory of Pattern Recognition Chinese Academy of Sciences). There are 108 classes or total number of Iris Images are 756. Fig.2 shows a sample of the image of an eye from this database

B. *Iris* S*egmentation*

The main motive behind Iris Segmentation is to remove the non-useful information like the sclera and the pupil information and extract the region of interest. First, the pupil is detected and then the Iris-sclera boundary is detected which is usually done in two steps.

B.1. *Detection of Pupillary Boundary*

As we all know that pupil is a very dark blob of a certain minimum size in the picture and no other segment of continuous dark pixels are of the same size. This algorithm finds the center of the pupil and two radial coefficients as the pupil is not always a perfect circle. To find the pupil, we first need to apply a linear threshold in the image,

$$g(x) = \begin{cases} f(x) > 70 : 1 \\ f(x) \leq 70 : 0 \end{cases}$$

Where *f(x)* is the original image and *g(x)* is the threshold image. Pixels with intensity greater than the empirical value of 70 (in a 0 to 255 scale) are dark pixels, therefore converted to 1 (black). Pixels smaller than or equal to 70 are assigned to 0 (white).Fig.3 shows the thresholded image of the pupil. Since, the eyelashes also satisfy the threshold condition, they are visible in the figure.

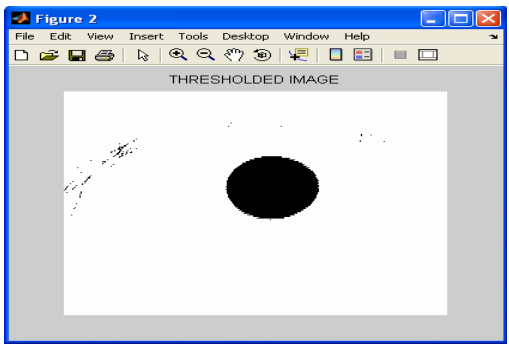

Fig.3 : Thresholded Image

To eliminate the area of eyelashes, search of a region of 8 connected pixels with value 1 is carried out. The report on CASIA database indicates that a value of 2500 is sufficient for pupil region. The eyelashes definitely have much smaller region than the pupil region, hence the area associated with eyelashes would be much smaller than 2500. Using this knowledge, one can cycle through all regions and apply the following condition:

*for each Region R*
*if AREA(R) < 2500*
*set all pixels of R to 0*

Thus the pupil is separated and the centroid $(x_{cp}, y_{cp})$ of the pupil is extracted. Also horizontal and vertical radius are calculated. Fig.4 shows the thresholded image in which the eyelashes has been cropped out by Freeman's Chain coding [3].

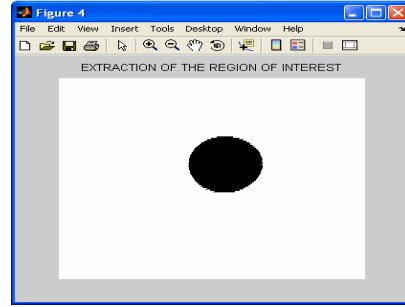

Fig.4 : Image with the Eyelashes Cropped out.

Fig.5 shows the pupil in the original eye image with centroid $(x_{cp}, y_{cp})$, horizontal radius $r_x$ and vertical radius $r_y$.

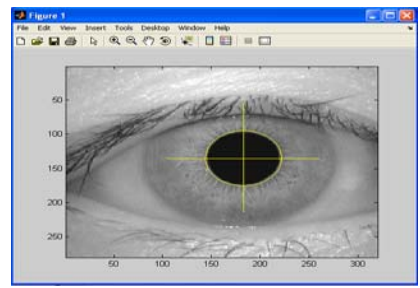

Fig. 5 : Detection of the centroid $(x_{cp}, y_{cp})$ and horizontal radius $r_x$ and vertical radius $r_y$

B.2. *Iris Edge Detection*

After detecting the pupil the next step is to find the contour of the Iris. Already, we have detected the pupil location and we have the knowledge that it is concentric to the outer perimeter of the Iris. But, the problem is that sometimes the eyelid may occlude part of the Iris. Also, the Iris center may not match with the pupil center, and we will have to deal with strips of Iris of different width around the pupil. This method takes into consideration the fact that areas of the Iris at the right and left side of the pupil contain the most significant information that is useful for data extraction. The areas above and bellow the pupil carry unique information, but it is very common that they are totally or partially occluded by eyelash or eyelid.

Fig.6 shows the steps how to find the right edge of the Iris as shown in the image. Yellow line passes through $y_{cp}$ (center of pupil) of original image. Red line shows the pixel intensities of that line.

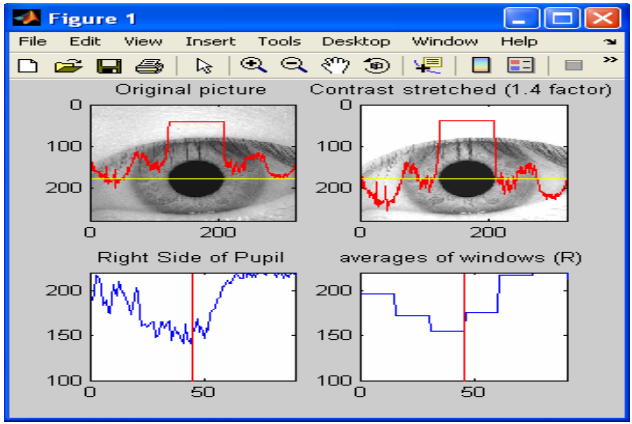

Fig.6: Iris Edge Detection

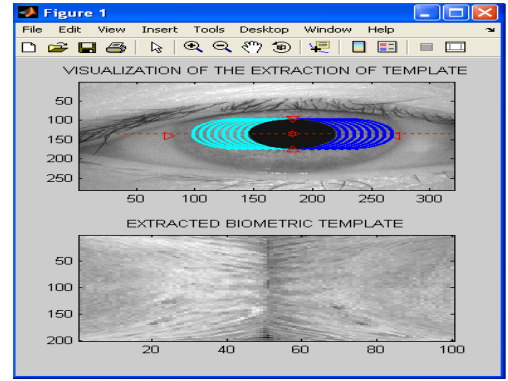

Fig.7 Extraction of the Biometric Template

The strategy adopted for Iris detection is to trace a horizontal imaginary line that crosses the whole image passing through the center of the pupil. Starting from the edges of the pupil, we analyze the signal composed by pixel intensity from the center of the image towards the border and try to detect abrupt increases of intensity level. Although the edge between the Iris disk and the sclera is most of the times smooth, it is known that it always have greater intensity than Iris pixels. We intensify this difference applying a linear contrast filter. It is possible that some pixels inside the Iris disk are very bright, causing a sudden rise in intensity. That could mislead the algorithm to detect that Iris edge at that point. To detect such edges, we take the average intensity of small windows when the sudden rises occur from these intervals.

C. *Feature Extraction*

Once the segmentation has been performed and the region of interest (ROI) has been extracted. The next step is to extract the feature so as to reduce the problem of dimensionality. The Iris Basis Images are extracted which are used as templates for training the neural network. Extraction of Iris Basis reduces the dimension as the non-useful information is cropped out. Pixels on either side of the pupil are collected and one reduced image of the Iris is formed. The main aim is to collect the desired number of Iris Basis rows and columns. We can better visualize this strategy by looking at Fig.7 as shown below. Up till now though the dimension has been reduced significantly it is too high for classification. So, singular value decomposition is employed for further reduction in dimension.

C.1. *Singular Value Decomposition (SVD)*

Singular Value Decomposition is a powerful tool for decomposing the Iris Basis matrix. SVD exposes the hidden geometry of the matrix. In this paper SVD is used as a dimensionality reduction tool. The basic operation of SVD relies on the factorization of an MxN matrix (M ≥ N) into three other matrices on the following form:

$$A = U_T \delta V$$

Where the superscript "*T*" denotes transpose. *U* is an MxM orthogonal matrix, *V* is an NxN orthogonal matrix and *δ* is an MxN diagonal matrix with $s_{ij}=0$ if i≠j and $s_{ii} \geq s_{i+1,i+1}$. The two important aspects to be noted here are:

1. *δ* is zero everywhere except in the main diagonal. This leads to reduction in the dimension of the input pattern from a matrix MxN to only a vector of N elements.
2. Only the first *k* elements contain substantial information, and the vector tail without significant loss of information can be cropped out.

Fig.8 shows a plot of SVD pattern vectors which very well depicts the second property mentioned above.

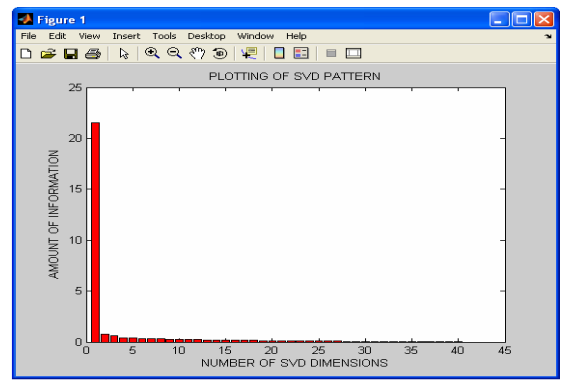

Fig.8: SVD Pattern Vectors.

## III. CLASSIFICATION

A Feed Forward Error Back-Propagation neural network is used for classification. Here, the first five patterns of total 7 patterns of each of the 108 classes of CASIA database are used for network training and the remaining two are used for network testing. Our network implements the classical 3-layer architecture: Input layer, Hidden layer and Output layer. The input layer contains as much neurons as the dimensionality of the pattern vector, which is 108 in present case. The number of neurons in the hidden layer is approximately double as that of input layer for good classification results.

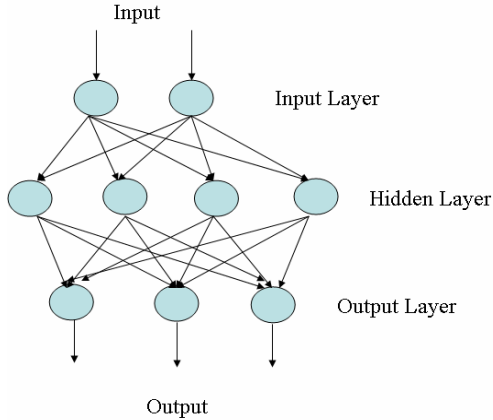

Fig.9 Architecture of Feed Forward Back propagation Neural Network

A. *Network Design*

Following are some parameters set for network training:
**Training function**: traingda (Adaptive learning rate)
**Initial learning rate**: 0.2
**Learning rate increment**: 1.05
**Epochs**: 50,000
**Error goal**: $5 \times 10^{-7}$
**Minimum gradient**: $1 \times 10^{-9}$

As mentioned earlier, the number of neurons in the output layer corresponds to the number of classes to recognize. When the network is trained in supervised mode, a target vector is also presented to the network. This target vector has every element set to zero, except on the position of the target class that will be set to 1. The idea behind this design decision is that for each input pattern X presented to the network, an output vector Y is produced. This vector has the number of elements equal to numbers of output neurons. Each output neuron implements a squashing function that produces a real number in the range [0, 1]. To determine which class is being indicated by the network, we select the maximum number in Y and set it to 1, while setting all other elements to zero. The element set to one indicates the classification of that input pattern. Fig.10 shows the convergence behavior of neural network using Gradient Descent algorithm for a typical case as obtained from MATLAB. It is seen that the MSE was reached within 2150 epochs and the performance goal was met giving 100% classification.

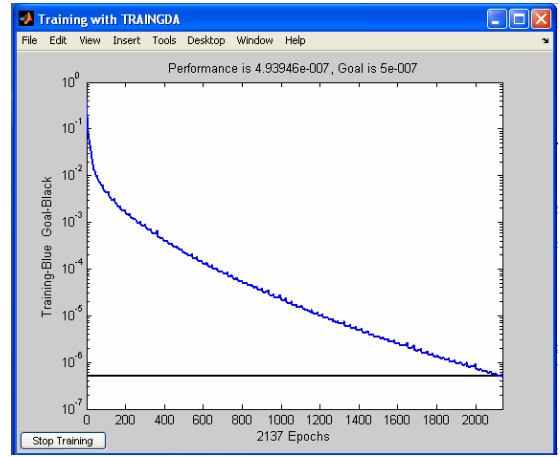

Fig.10 Training of the Feed Forward Neural Network

The various experiments performed to train the neural network and test the Iris pattern include Iris-basis images of 40*40 pixels quantized from the original Iris image with a mask of 3*3 pixels. The SVD algorithm was developed to output the vectors with 3, 10, 20 and 40 dimensions. The target classes for classification were varied from 3 to 50 as shown in Table 1.

While simulating using MATLAB it was found that as the number of classes and cases increased, the network had more difficulty in learning the proper discriminatory weights. The network was able to reach the MSE goal within the specified number of epochs. As for number of classes greater than 6, the MSE goal was not attained anymore, but the MSE kept decreasing until the maximum number of epochs was reached. We feel that increasing the number of epochs may allow the network to eventually converge. However, this may not be justified.

TABLE I. CLASSIFICATION OF DIFFERENT CLASSES USING DIFFERENT DIMENSION

| Number of Classes | Classification Rate | | | |
|---|---|---|---|---|
| | Number of Dimensions | | | |
| | 3D | 10D | 20D | 40D |
| **3** | 50% | 100% | 100% | 100% |
| **4** | 50% | 87.5% | 100% | 100% |
| **5** | 50% | 80% | 100% | 100% |
| **6** | 41.66% | 58.33% | 91.67% | 91.67% |
| **7** | 57.14% | 64.29% | 92.86% | 78.57% |
| **8** | 43.75% | 62.5% | 87.5% | 68.75% |
| **9** | 61.11% | 55.55% | 94.44% | 83.33% |
| **10** | 35% | 55% | 70% | 65% |
| **20** | 27.5% | 50% | 55% | 52.5% |
| **40** | 2.5% | 2.5% | 2.5% | 2.5% |
| **50** | 2% | 2% | 2% | 1% |

## IV. RESULTS

Table I shows the classification rate when the number of SVD dimensions and number of classes are varied. Since, only the first few dimensions contains the substantial information it can be seen that increasing the SVD dimensions after 20 D does not make much of a difference in the Classification rate. But, it can be seen that as the number of classes are increased beyond 9 the system tends to become over biased towards one class and the classification rates becomes poor. This behavior is also as shown in Fig. 11.

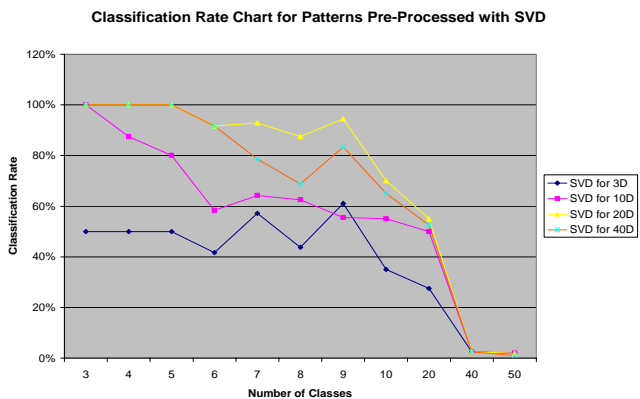

Fig.11 Classification Rate Chart for patterns Pre-Processed with SVD

It is seen that the classification rate drops abruptly to very low value for classes in excess of 20 and is independent of SVD dimension. So, this method of classification perhaps cannot be applied for classes above 20.

## V. CONCLUSION

A method for pattern recognition based on Singular Value Decomposition (Feature Extraction) with Error-Back Propagation method of Neural Network (Recognition) was successfully implemented for Iris recognition. Number of experiments was carried out with varying SVD dimensions and number of classes. The study results of our work indicates that optimum classification values are obtained with SVD dimension of 20 and maximum number of classes as 9. For SVD classes around 20 the performance of the network drops abruptly to around 2% and becomes independent of SVD dimension. In such cases the network becomes biased towards one class and is not recommended for classification.

## ACKNOWLEDGEMENT

This research paper uses the CASIA Iris image database collected by Institute of Automation, Chinese Academy of Sciences.

AUTHORS PROFILE

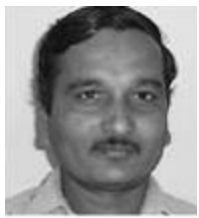

**Mr. Babasaheb G. Patil** : He received his M.E. Electronics degree in 1990 and B.E. Electronics in 1988. He is currently working as a associate professor in department of Electronics in Walchand College of Engineering, Sangli, Maharashtra, India. He is having keen interest in image processing and communication. He is carrying out research work in the field of Image Processing.

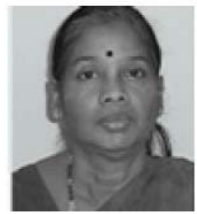

**Dr. (Mrs) Shaila Subbaraman :** She received M-Tech degree from IISc. Bangalore in 1975 and Ph.D. from IIT Bombay in 1999. She worked in Semiconductor Device Manufacturing company from 1975 to 1989. Currently she is Professor in Department of Electronics in Walchand College of Engineering, Sangli, Maharashtra, India. She has keen interest in the field of Microelectronics and VLSI Design.